\documentclass[10pt,twocolumn,letterpaper]{article}
\usepackage{scrextend}
\usepackage{authblk} 
\deffootnote{1.5em}{0em}{\textsuperscript{\textasteriskcentered}} 
\usepackage{authblk}
\usepackage[normalem]{ulem}
\usepackage{iccv}    

\usepackage{xcolor}
\usepackage{color}
\usepackage{url}
\usepackage{multirow}
\usepackage{epsfig}
\usepackage{threeparttable}
\usepackage{algorithmicx,algorithm}
\usepackage{algpseudocode}
\usepackage{bbm}
\usepackage{multicol}
\usepackage{makecell}
\usepackage{graphicx}
\usepackage{colortbl}
\usepackage{booktabs}

\definecolor{iccvblue}{rgb}{0.21,0.49,0.74}
\usepackage[pagebackref,breaklinks,colorlinks,allcolors=iccvblue]{hyperref}

\def\paperID{14} 
\def\confName{ICCV workshop}
\def\confYear{2025}

\title{Spoofing-aware Prompt Learning for Unified Physical-Digital \\ Facial Attack Detection}

\author{Jiabao Guo$^1$\quad Yadian Wang$^1$\quad Hui Ma$^2$\quad Yuhao Fu$^1$\quad Ju Jia$^3$\quad Hui Liu$^4$\quad Shengeng Tang$^1$\quad \quad \quad Lechao Cheng$^1$\quad Yunfeng Diao$^1$\quad Ajian Liu$^{5*}$ \\
$^1$HFUT, China\quad $^2$MUST, China \quad $^3$SEU, China \quad $^4$CCNU, China\quad $^4$CASIA, China\\ 
{\tt\small \{garbo\_guo,tangsg,chenglc,diaoyunfeng\}@hfut.edu.cn}\\
{\small
\texttt{jiaju@seu.edu.cn} \quad \texttt{liuh@ccnu.edu.cn} \quad \texttt{ajian.liu@ia.ac.cn}}}

\begin{document}
\maketitle
\footnotetext{Corresponding author}
\begin{abstract}
Real-world face recognition systems are vulnerable to both physical presentation attacks (PAs) and digital forgery attacks (DFs). We aim to achieve comprehensive protection of biometric data by implementing a unified physical-digital defense framework with advanced detection. 
Existing approaches primarily employ CLIP with regularization constraints to enhance model generalization across both tasks. However, these methods suffer from conflicting optimization directions between physical and digital attack detection under same category prompt spaces. To overcome  this limitation, we propose a \textbf{S}poofing-aware \textbf{P}rompt \textbf{L}earning for \textbf{U}nified \textbf{A}ttack \textbf{D}etection (\textbf{SPL-UAD}) framework, which decouples optimization branches for physical and digital attacks in the prompt space. Specifically, we construct a learnable parallel prompt branch enhanced with adaptive Spoofing Context Prompt Generation, enabling independent control of optimization for each attack type. Furthermore, we design a Cues-awareness Augmentation that leverages the dual-prompt mechanism to generate challenging sample mining tasks on data, significantly enhancing the model's robustness against unseen attack types.
Extensive experiments on the large-scale UniAttackDataPlus dataset demonstrate that the proposed method achieves significant performance improvements in unified attack detection tasks.

\end{abstract}

\section{Introduction}
\label{sec:intro}

Facial recognition systems continue to be vulnerable to various attack methods, primarily categorized into physical and digital attacks. Physical attacks~\cite{liu2021face,liu2019multi,liu2021casia,liu2022contrastive} typically include print attacks, replay attacks, and mask attacks, while digital attacks~\cite{diao2025tasar,MSGM,diao2021basar,jia2022consensus} are often challenged by various generation technologies~\cite{tang2025discrete,guo2025shaping,tang2025sign,wang2025text2lip}. Research on detecting physical attacks commonly focuses on developing specialized networks capable of automatically extracting spoofing indicators and deceptive characteristics from multimodal data sources. In contrast, digital attack detection approaches often utilize frequency-domain information and facial action unit relationships to differentiate authentic faces from manipulated ones. However, existing detection methods generally lack effectiveness across diverse attack types and categories for Unified Attack Detection (UAD).

Existing unified detectors often adapt CLIP via regularization or prompt tuning to enhance cross-task generalization, yet they inherit several fundamental limitations~\cite{zhang2025patfinger,jia2025prompt}. Previous approaches learn common spoofing representations by aligning distributions across heterogeneous attacks~\cite{jia2020single, liu2022contrastive}, which inevitably suppresses physical-specific or digital-specific signals and distorts semantic structure, yielding over-smoothed, weakly discriminative features. CLIP-based methods seek to counter this by using textual features as adaptive classifiers~\cite{radford2021learning} and by replacing hand-crafted templates with learnable contexts (e.g., CoOp positions a learnable context with a trailing [CLASS] token. CoCoOp adds an conditional token via a lightweight meta-net)~\cite{zhou2022coop, zhou2022cocoop}. However, two obstacles persist for UAD. First, semantic misalignment: category tokens like “Real”/“Fake” provide little descriptive value to CLIP’s text encoder, which is trained to ground rich natural language, not binary labels. This weak supervision undermines zero-shot transfer and causes ambiguous text–image alignment ~\cite{radford2021learning}. Moreover, UAD samples are semantically multi-explained (e.g., “This is a Real face” vs. “This is a Fake face”), but such minimal phrases are largely uninformative to CLIP. Second, optimization conflict and prompt rigidity. A single shared prompt space entangles gradients from physical presentation attacks and digital forgeries, blurring attack-specific cues and causing negative transfer. A single fixed prompt cannot capture dataset and device cues (like sensor noise, lighting, and shooting distance) or the maker-specific artifacts of modern forgeries, which leads to weak generalization to unseen attacks and new domains.

These observations motivate a spoofing-aware, decoupled prompt learning strategy for unified detection. Rather than collapsing all attacks into a single latent criterion, we aim to construct semantically rich, spoof-aware contexts that better align with CLIP’s language priors while preserving the downstream backbone. Concretely, we seek: (1) explicit decoupling of physical and digital branches within the prompt space to avoid gradient interference and retain attack-specific evidence; (2) data-driven context construction that leverages class-level embedding structure (e.g., clustering centers with lightweight linear projections) to produce multi-granularity textual and visual pre-contexts; and (3) cues-awareness augmentation that exploits dual prompts to mine hard examples and stress-test the decision boundary. This direction promises lightweight adaptation with minimal parameter overhead while substantially improving robustness and transferability across sensors, environments, and unseen generative attacks.

In this work, we propose a novel spoofing-aware prompt learning framework for unified attack detection (SPL-UAD). The central idea of SPL-UAD is to exploit spoof-aware context to improve the task-awareness of learnable prompts. To generate spoof-aware context prompt, we conduct clustering on the class embeddings of a downstream task and perform linear transformation on the clustering centers to yield spoof-aware context, which leads to a spoof-aware context generalization module that can keep the backbone structure of the downstream task while being lightweight. Then, the spoof-aware context prompt is combined with the learnable prompt tokens and their interaction through the well-pretrained encoders is exploited to reinforce the task-awareness of learnable prompts. To suppress the overfitting induced by the scarcity of task-specific data, the prompted class and visual embeddings are encouraged to be consistent with their CLIP peers. Our method maintains simplicity in design and improves the task-awareness of the resulting prompts, thus yielding competitive results.

In summary, our spoofing-aware prompt learning framework has the following main contributions:

\begin{itemize}
\item We propose a spoofing-aware prompt learning framework that constructs spoof-aware contexts to overcome semantic limitations of naïve CLIP prompting in UAD. 

\item We decouple physical and digital optimization within the prompt space and propose cues-awareness augmentation to mine hard cases and enhance robustness.

\item We demonstrate the superiority of our SPL-UAD in the UniAttackDataPlus.

\end{itemize}

\section{Related Work}
\label{sec:relatedwork}

\subsection{Physical Attack Detection}

Physical Attack Detection (PAD) distinguishes real faces from physical spoofs (e.g., prints, replays, 3D masks).  With deep learning, CNNs dominated, initially treating PAD as binary classification~\cite{li2016original}. Later works added auxiliary supervision for intrinsic physical differences (e.g., pseudo depth, reflection, texture maps) to learn live-spoof distinctions. Disentangled learning like PIFAS~\cite{liu2022disentangling} and adversarial frameworks like AA-FAS~\cite{liu2023attack} further improved representation. Despite intra-dataset success, performance degrades under domain shifts, addressed by Domain Generalization (DG) techniques aligning cross-source features without target data—including adversarial alignment SSDG~\cite{jia2020single}, unified transition modeling SA-FAS~\cite{sun2023rethinking}, instance alignment IADG~\cite{zhou2023instance}. Recent Vision-Language Models (VLMs) like CFPL-FAS~\cite{liu2024cfpl}, S-CPTL~\cite{guo2024style}, and CCPE~\cite{guo2025domain} use CLIP/LLM features for robust generalization. Flexible-modal frameworks (e.g., FM-ViT~\cite{liu2023fm}, FM-CLIP~\cite{liu2024fm}) and innovations like source-free adaptation~\cite{li2025optimal}, and multimodal alignment mmFAS~\cite{chen2025mmfas} further advance the field. Critical to these advances are large-scale datasets and benchmarks. The CASIA-SURF series~\cite{liu2019multi,liu2021casia,liu2021cross} established multi-modal benchmarks (RGB, Depth, IR) with explicit ethnic labels, enabling rigorous evaluation of cross-ethnicity generalization. To address high-fidelity 3D mask threats, HiFiMask~\cite{liu2022contrastive,liu20213d} introduced 54,600 videos spanning 75 subjects and 225 masks across 7 sensors, catalyzing research via organized challenges.
However, most methods remain limited by modality dependencies or artifact overfitting, prompting demand for unified frameworks beyond physical paradigms.
\subsection{Digital Attack Detection}
Digital forgery detection targets pixel-level manipulations that threaten face recognition security, including deep synthesis, identity swaps, and attribute editing. Early approaches framed the task as binary classification over spatial artifacts, followed by methods that exploited frequency inconsistencies, blending boundaries, and resolution mismatches~\cite{chen2022self,li2018exposing,qian2020thinking}. The standard pipeline fine-tunes CNNs on authentic and manipulated faces ~\cite{chen2022self,dang2020detection,rossler2019faceforensics++,wang2019fakespotter}, but has shown limited transfer across generators and datasets. Recent trends emphasize generalization and multimodal integration. Vision–language models such as FFTG~\cite{sun2025towards}, Forensics Adapter~\cite{cui2025forensics}, and VLF-FFD ~\cite{peng2025mllm} mitigate annotation hallucination and adapt CLIP to forensic cues. Continual and incremental paradigms, including SUR and LID~\cite{cheng2025stacking} and HDP with UAP~\cite{sun2025continual}, reduce catastrophic forgetting via feature isolation and staged updates. Architecturally, specialized designs like MFCLIP~\cite{zhang2025mfclip} align image–noise–language spaces to better expose diffusion forgeries; distilled transformers model local–global artifacts. Wavelet-CLIP~\cite{baru2025wavelet} improves cross-dataset transfer through frequency-aware alignment. Bias and semantics-oriented solutions, including FairFD ~\cite{liu2025thinking}, semantic redefinition~\cite{zou2025semantic}, and distributional learning with FakeDiffer~\cite{wang2025fakediffer}, aim to reduce spurious correlations and close the semantic gap. Test-time techniques further enhance reliability, from spatial–frequency prompting~\cite{duan2025test} to explainable detection in M2F2-Det~\cite{guo2025rethinking}. Overall, the field is transitioning from artifact-specific classifiers toward adaptable, bias-aware, and multimodal systems, yet open-set robustness under unseen generators and real-world degradations remains a central challenge.

\subsection{Unified Face Attack Detection}
Recent research has shifted toward developing integrated frameworks capable of simultaneously detecting both physical presentation attacks (PAs) and digital forgery attacks (DFs) within a unified model~\cite{he2024joint}. This paradigm addresses the inefficiency of maintaining separate detection systems while improving the robustness against hybrid threats. Pioneering datasets have been developed to support this integration. Grandfake~\cite{deb2023unified} and JFSFDB~\cite{yu2024benchmarking} combined existing PAD and FFD datasets, while UniAttackData~\cite{fang2024unified} introduced identity-consistent samples where each subject includes all attack types, reducing bias from non-critical factors. UniAttackDataPlus~\cite{liu2025benchmarking} addresses the limitations of outdated digital attack samples and insufficient coverage of emerging generative forgeries in existing datasets by hierarchically integrating modern attacks. 

Methodologically, Yu et al.~\cite{yu2024benchmarking} established the first joint benchmark using both visual appearance and physiological rPPG cues, enhancing periodicity discrimination through spatio-temporal signal maps and wavelet transforms. To mitigate modality bias, they implemented weighted normalization before fusion, demonstrating that joint training improves generalization across spoofing and forgery tasks. Fang et al.~\cite{fang2024unified} leveraged Vision-Language Models (VLMs) with their UniAttackDetection framework, incorporating Teacher-Student Prompts for unified/specific knowledge and Unified Knowledge Mining for comprehensive feature spaces. HiPTune~\cite{liu2025benchmarking}, which adaptively selects semantic-space criteria via Visual Prompt Trees. Key architectural advances focus on handling feature distribution divergence. La-SoftMoE~\cite{zou2024softmoe} employed re-weighted Mixture-of-Experts (MoE) with linear attention to process sparse feature regions, while MoAE-CR~\cite{chen2025mixture} introduced class-aware regularization via Disentanglement and Cluster Distillation Modules to enhance inter-class separability. SUEDE~\cite{xie2025suede} combined shared experts (common features) and routed experts (attack-specific features) with CLIP alignment. Reconstruction-based approaches also gained traction. Cao et al.~\cite{cao2024towards} proposed dual-space reconstruction in spatial/frequency domains to model genuine face fundamentals, filtering redundant information to isolate attacks as outliers. Despite progress, most methods rely on single classification criteria that struggle with advanced attacks. Current limitations include insufficient exploration of ID-consistent learning and under-use of physiological signals, highlighting promising directions for future work.

\section{Methodology}
\subsection{Preliminaries}

Contrastive Language-Image Pre-training (CLIP)~\cite{radford2021learning} achieves strong visual representation learning through its dual-encoder architecture that separately processes images and text. Formally, we define a dataset $\mathbb{D}={(x_i, t_i)}_{i=1}^{B}$ containing $B$ image-text pairs across $C$ categories. The image modality is represented by $x_i \in \mathbb{R}^{H \times W \times 3}$, and $t_i$ provides the corresponding text description in natural language form. Patch-based feature extraction begins with dividing the image into $N_v$ regular grids, producing patch embeddings $\boldsymbol{E}_i^p \in \mathbb{R}^{N_v \times d}$ via embedding projection. These embeddings then undergo Transformer to produce the final image representation $v_i \in \mathbb{R}^{d}$. Given category names $\left\{\left[\text{\textit{CLASS}}\right]_c\right\}_{c=1}^C$ in the dataset, each class-specific prompt is formulated as $t^{clip}_c = \left\{\text{\textit{A  photo of a} } \left[\text{\textit{CLASS}}\right]_c\right\} $ for CLIP adaptation. The tokenized words are first mapped to embedding space, then pass through transformer layers to produce textual features $l^ {clip}_c\in \mathbb{R}^d$. Prediction probabilities are computed using both visual features $\boldsymbol{v}{i}$ and textual embeddings via cross-modal similarity:
\begin{equation}
p(y=c \mid v_i) = \frac{ \exp\left( \frac{\cos(v_i, l^{\text{clip}}_y)}{\tau} \right) }{ \sum_{c=1}^{C} \exp\left( \frac{\cos(v_i, l^{\text{clip}}_c)}{\tau} \right) }
\end{equation} where $\operatorname{\cos}(\cdot,\cdot)$ refers specifically to cosine similarity and $\tau$ is a temperature parameter. Prompt learning in CLIP enables task-specific adaptation through hand-crafted templates while maintaining the frozen status of both image and text encoders during downstream transfer. 
 
\subsection{Overview}
\begin{figure*}[ht]
\centering
\includegraphics[scale=0.6]{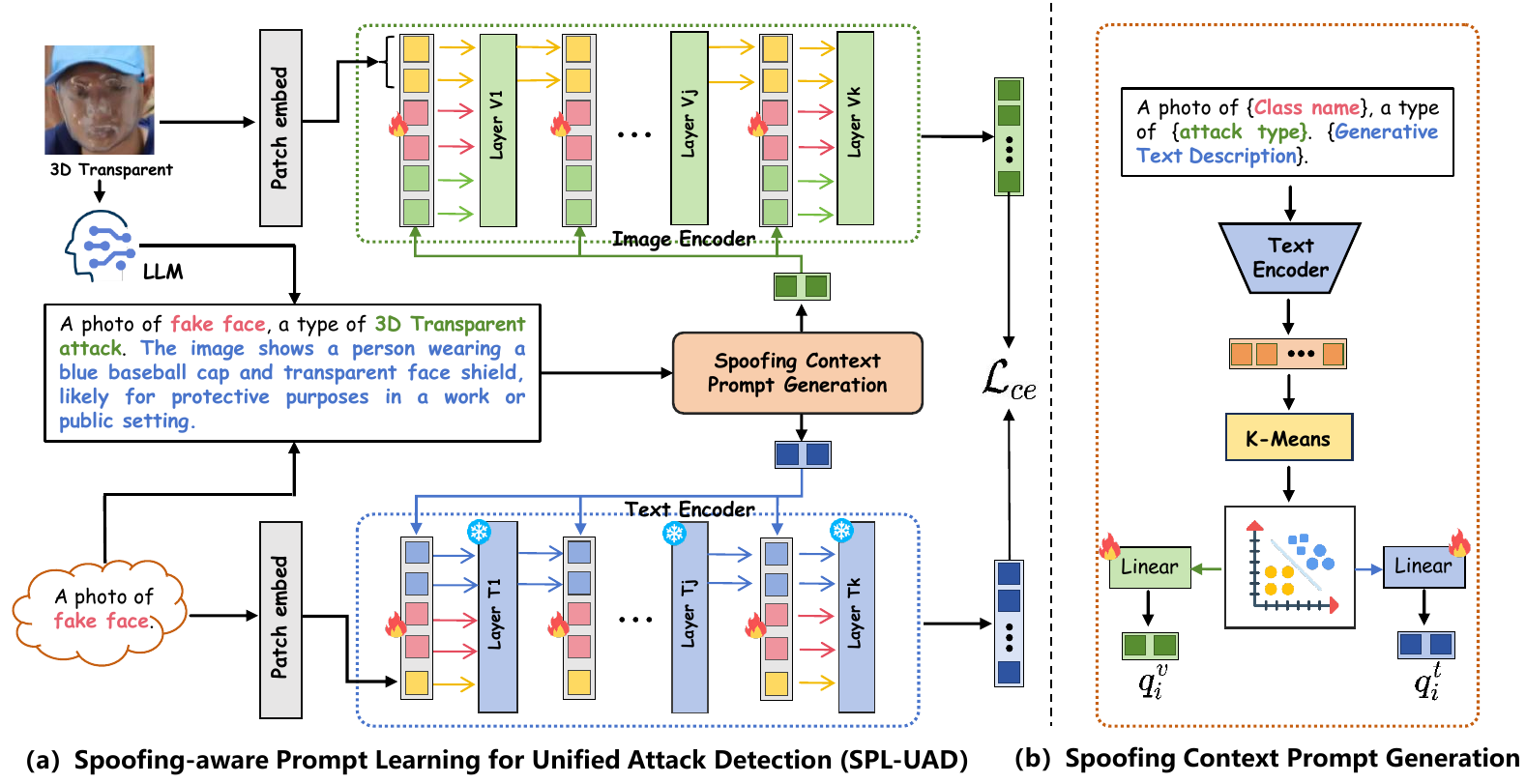}
\caption{Overview of the proposed SPL-UAD framework. (a) Spoofing-aware Prompt Learning. An image is tokenized into patch embeddings and processed by the frozen CLIP iamge encoder, while text tokens are fed into the frozen text encoder. We inject learnable prompts together with spoof-aware context at multiple transformer layers. A dual-branch design decouples optimization for physical and digital attacks, mitigating conflicting gradients and preserving attack-specific cues. Cross-modal similarities between the resulting visual and textual features are used for classification, and representations are further organized by K-Means to support context construction. (b) Spoofing Context Prompt Generation (SCPG). We cluster class-level embeddings to obtain centers and apply lightweight linear projections to yield textual and visual context that align with encoder hidden sizes. Multi-Granularity Spoof-Aware descriptions enrich semantics for both real and spoof classes. The combined design provides informative pre-context, promotes stable text–image interactions, and enables cues-awareness augmentation to mine hard examples, ultimately improving robustness to both physical and digital attacks.}
\label{The overview of SPL-UAD}
\end{figure*}

Fig.~\ref{The overview of SPL-UAD} presents an overall architecture of our proposed Spoofing-aware Prompt Learning framework for Unified Face Attack Detection (SPL-UAD). Unlike prior works that endeavor to design category-based prompts, SPL-UAD proposes a spoofing-aware prompting approach where the problem of category names (i.e. real/spoof) without semantics can be solved while the diverse characteristics of categories are guaranteed. The framework first generates spoofing context prompts. Visual inputs are processed via a patch embedding module, while textual inputs pass through a word embedding module following a predefined template. These embeddings are then passed through multiple transformer layers of visual and textual encoders. The resulting features from these encoders are fed into linear layers and combined within a CLIP module. Finally, the obtained representations are clustered using K-Means to support spoofing detection tasks.

\subsection{Spoofing Context Prompt Generation}

Class embeddings derived from the CLIP text encoder are task-aware and serve as suitable contextual priors. To construct spoof-aware textual and visual contexts while preserving the backbone structure of class embeddings, we apply K-Means to $\{w_c^{\text{CLIP}}\}_{c=1}^{C}$, obtaining $K$ cluster centers $\{q_i\}_{i=1}^{K}$. The spoof-aware textual and visual context prompts are then defined as:
\begin{equation}
q_i^{t} = W^{t} q_i,
\end{equation}
\begin{equation}
q_i^{v} = W^{v} q_i,
\end{equation}
where $W^{t}$ and $W^{v}$ denote linear transformations for the text encoder and image encoder, respectively, with dimensions aligned to the internal architectures of their corresponding encoders. Consequently, the dimensions of $q_i^{t}$ and $q_i^{v}$ explicitly match the hidden sizes of the text and image encoders. This design requires optimizing only linear projection weights to generate spoof-aware contexts, resulting in a lightweight and computationally efficient framework with minimal parameter overhead.

\subsection{Cues-awareness Augmentation}
\begin{figure*}[h]
\centering
\includegraphics[scale=0.9]{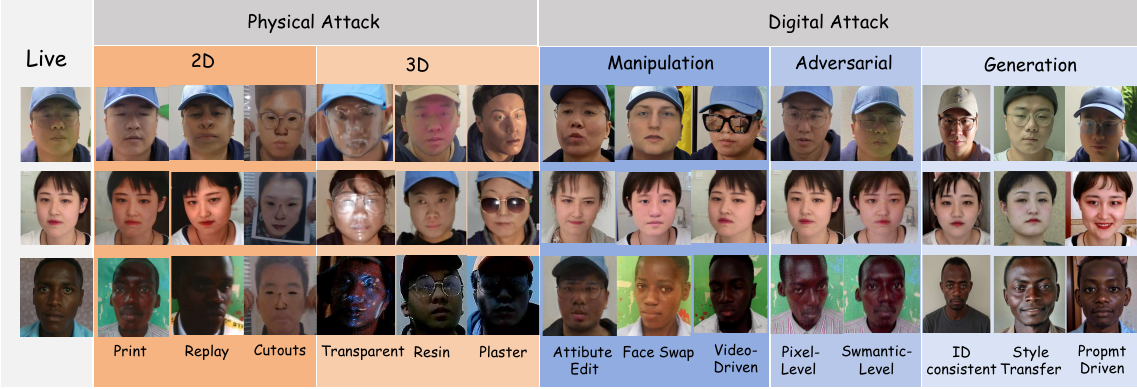}
\caption{The Samples of UniAttackDataPlus Dataset~\cite{liu2025benchmarking}. Samples from UniAttackDataPlus covering both physical and digital attack families. Physical attacks include 2D prints, replay videos, cutouts, and diverse 3D masks such as transparent shields, resin, and plaster, collected under varied sensors and environments. Digital attacks span pixel-level manipulations and semantic-level edits, with ID-consistent pairing to real subjects. The breadth of capture conditions and the hierarchical taxonomy of attack types encourage models to focus on spoof-specific cues rather than incidental correlations, enabling unified evaluation across diverse threats.}
\label{fig2}
\end{figure*}

After acquiring textual and visual contexts, we enhance the task-awareness of learnable prompts by allowing them to interact with their respective contexts through well-trained text and image encoders. Specifically, assuming the image/text encoder consists of 
$L$ transformer blocks, we construct the input embeddings $\mathcal{S}_l^{t}$ for the $l$-th transformer block of the text encoder as follows:

\begin{equation}
\mathcal{S}_l^{t}
= \left[ t _ { l } ^ { s o s } , q _ { 1 } ^ { t } , \cdots , q _ { K } ^ { t } , p _ { l_1 } ^ { t } , \cdots ,  p _ { l_{M ^  t } } ^ { t } , c _ { l_i } , t _ { l } ^ { c o s } \right] 
\end{equation}
Where $ t _ { l } ^ { s o s }$ is the start token input to the $ l $-th transformer block, $ t _ { l } ^ { e o s } $ is the end token input to the $ l $-th transformer block, $ \left\{ p _ { l_i } ^ { t } \right\} _ { i = 1 } ^ { M^t }$ is the set of learnable textual prompts input to the $ l $-th transformer block with a quantity of $M^t$, and $ c _ { l_i }$ is the word embedding(s) input to the $ l $-th transformer block. It is worth noting that $  t _ { l } ^ { s o s } , t _ { l } ^ { e o s } $ , and $ c _ { l_i }$ are the outputs of the previous transformer block. Similarly, we construct the input embeddings to the $ l $-th transformer block of the image encoder as:

\begin{equation}
\mathcal{S}_l^{v}
= \left[ e _ { l } ^ { c l s } , x _ { l_1 } ^ { p } , \cdots , x _ { l_M } ^ { p } , q _ { 1 } ^ { v } , \cdots , q _ { K } ^ { v } , p _ { l_1 } ^ { v } , \cdots , p _ { l_{M ^ v } } ^ { v } \right] 
\end{equation}
 where $e_{l}^{\mathrm{cls}}$ is the class token fed to the $l$-th transformer block, $\{x_{l_m}^{p}\}_{m=1}^{M}$ are the $M$ patch embeddings input to the same block, and $\{p_{l_i}^{v}\}_{i=1}^{M^v}$ are the $M^v$ learnable visual prompts associated with that block. Importantly, $e_{l}^{\mathrm{cls}}$ and $\{x_{l_m}^{p}\}_{m=1}^{M}$ are propagated from the previous block, while $\{p_{l_i}^{v}\}_{i=1}^{M^v}$ and $\{q_{i}^{v}\}_{i=1}^{K}$ are independently parameterized and do not depend on outputs from earlier blocks.

\section{Experiments}
\subsection{Experimental Settings}

{\flushleft \textbf{Datasets}}
UniAttackDataPlus~\cite{liu2025benchmarking} is a comprehensive benchmark dataset specifically designed for training and evaluating facial attack detection systems, serving as the largest publicly accessible unified dataset in this area. The dataset comprises 18,250 authentic videos from three ethnic populations (African, East Asian, and Central Asian), each subjected to 54 different attack scenarios (14 physical, 40 digital) under diverse lighting conditions, backgrounds, and acquisition devices, samples can be found in Fig.~\ref{fig2}. The diverse acquisition backgrounds of authentic faces provide a more varied dataset, encouraging models to focus on discriminative features rather than irrelevant factors. We utilize a sub-dataset of UniAttackDataPlus, comprising 21528 training samples and 5383 testing samples.

\begin{table*}[h]
    \centering
     \caption{Results on UniAttackDataPlus.}
     \label{tab:overview_results}
    \begin{tabular}{ >{\arraybackslash}p{4.5 cm}
  >{\centering\arraybackslash}p{2 cm}
  >{\centering\arraybackslash}p{2 cm}
  >{\centering\arraybackslash}p{2 cm}
  >{\centering\arraybackslash}p{2 cm}}
        \toprule
        \textbf{Method} & \textbf{ACC (\%)}  & \textbf{AUC (\%)} &\textbf{EER (\%)}  & \textbf{ACER (\%)}  \\
         \midrule
        CLIP-V (PMLR'21)~\cite{radford2021learning}& 58.08  &43.10   & 49.87   &  49.99\\
        CLIP (PMLR'21)~\cite{radford2021learning}& 59.90  & 58.21    & 34.37   & 34.87 \\
        CoOp (IJCV'22)~\cite{zhou2022coop}& 27.18  & 21.97    &  71.29 & 71.29 \\
        CoCoOp (CVPR'22)~\cite{zhou2022cocoop}& 29.87  & 20.29  & 69.67 & 69.67   \\
        CFPL-FAS (CVPR'24)~\cite{liu2024cfpl} & 71.46 & 73.50 & 29.60 & 29.60 \\
        \textbf{SPL-UAD (Ours)} & 67.97 & 72.55 & 34.00 & 28.09 \\
        \midrule
    \end{tabular}
\end{table*}

\begin{table*}[h]
\centering
\caption{Ablation on SCPG and CAA. \checkmark indicates the module is enabled.}
\begin{tabular}{ >{\centering\arraybackslash}p{2 cm}
  >{\centering\arraybackslash}p{2 cm}
  >{\centering\arraybackslash}p{2 cm}
  >{\centering\arraybackslash}p{2 cm}
  >{\centering\arraybackslash}p{2 cm}
  >{\centering\arraybackslash}p{2 cm}}
\toprule
\textbf{SCPG} & \textbf{CAA} & \textbf{ACC (\%)} & \textbf{AUC (\%)} & \textbf{EER (\%)} & \textbf{ACER (\%)} \\
\midrule
& & 61.20 & 66.10 & 39.80 & 34.90 \\
\checkmark & & 65.40 & 70.80 & 36.70 & 31.20 \\
& \checkmark & 63.10 & 68.20 & 35.50 & 30.40 \\
\checkmark & \checkmark & 67.97 & 72.55 & 34.00 & 28.09 \\
\bottomrule
\end{tabular}
\label{tab:ablation_scpg_caa_switch}
\end{table*}

{\flushleft \textbf{Implementation Details.}}
To effectively demonstrate the advantages of our approach, we compare it against a range of existing methods using diverse network backbones. All facial images are pre-processed to a resolution of $224 \times 224 \times 3$ and divided into patches of size $14 \times 14$. The image and text encoders are adapted from the pre-trained ViT-B/16 model in CLIP, and each feature vector extracted has a dimensionality of 512. Our implementation is built on PyTorch, with training optimized using the Adam optimizer. The training process begins with an initial learning rate of $10^{-6}$ and employs a batch size of 32.

{\flushleft \textbf{Evaluation Metrics.}}
We assess our approach on a unified benchmark that spans both physical and digital attack settings. To enable comprehensive evaluation, we report the following standard metrics: Average Classification Error Rate (ACER), Area Under the ROC Curve (AUC), Accuracy (ACC) and Equal Error Rate (EER).

\subsection{Comparisons to Prior SOTA Results}
To illustrate our model's ability to adapt to the unified physical-digital
facial attack detection task, we give the result summarized in Tab.~\ref{tab:overview_results}.The table compares various methods, including conventional approaches for direct manipulation of image features and CLIP-based prompt engineering techniques, evaluated on the UniAttackDataPlus dataset. Our method, SPL-UAD, demonstrates strong performance across key metrics. Specifically, SPL-UAD achieves an ACER of 28.09\%, outperforming all other methods. In terms of EER, SPL-UAD achieves 34.00\%, slightly trailing behind CFPL-FAS but significantly outperforming other baselines such as CoOp and CoCoOp. For the AUC metric, SPL-UAD achieves 72.55\%, closely following CFPL-FAS, which leads in this aspect. Overall, SPL-UAD exhibits competitive performance across most metrics, highlighting its effectiveness in adapting to attack detection tasks.

\subsection{Ablation Studies}
We ablate the key components of SPL-UAD on UniAttackDataPlus. Removing SCPG and reverting to vanilla learnable prompts lowers ACC and AUC, showing that clustering with linear projection supplies meaningful context for stronger text–image alignment. Using only textual or only visual context provides complementary gains, and applying both together performs best. Disabling CAA increases EER and ACER, indicating that dual-prompt hard-sample mining enhances robustness to unseen attacks. A moderate augmentation intensity yields the best trade-off, while overly strong augmentation slightly reduces ACC due to distribution shift. From the table, adding SCPG to the baseline raises ACC from 61.20\% to 65.40\% and AUC from 66.10\% to 70.80\%. Adding CAA alone delivers larger improvements on robustness-oriented metrics, reducing EER from 39.80\% to 35.50\% and ACER from 34.90\% to 30.40\%. The combination of SCPG and CAA achieves the strongest overall result, reaching 67.97\% ACC and 72.55\% AUC while further lowering EER and ACER. In summary, SCPG mainly boosts discriminability as reflected by higher ACC and AUC, whereas CAA primarily strengthens robustness as seen in lower EER and ACER. The two modules reinforce each other.

\section{Conclusion}
This work tackles the critical challenge of unified physical and digital attack detection to enhance robust face recognition security. Existing CLIP-based approaches often face limitations due to conflicting optimization directions when utilizing shared prompt spaces for both attack types. To address this issue, we introduce the Spoofing-aware Prompt Learning for Unified Attack Detection (SPL-UAD) framework. The core innovation of SPL-UAD lies in decoupling the optimization pathways within the prompt space. This is achieved through: (1) a learnable parallel prompt branch with an adaptive Spoofing Context Prompt Generation module, which separately guides the optimization for physical and digital attacks, and (2) a Cues-Awareness Augmentation strategy that leverages the dual-prompt mechanism to generate challenging samples, improving robustness against unseen attacks. Comprehensive evaluations on the large-scale UniAttackDataPlus dataset demonstrate that SPL-UAD delivers significant performance improvements in unified attack detection, offering a more effective and reliable solution for comprehensive biometric security against diverse spoofing threats.





{
    \small
    \bibliographystyle{ieeenat_fullname}
    \bibliography{main}
}

\end{document}